\documentclass[sigconf, authorversion]{acmart}
\title[An Eye for an AI]{An Eye for an AI: Evaluating GPT-4o's Visual Perception Skills and Geometric Reasoning Skills Using Computer Graphics Questions}

\author{Tony Haoran Feng}
\orcid{0009-0005-8608-6961}
\affiliation{
  \institution{University of Auckland}
  \city{Auckland}
  \country{New Zealand}
}
\email{hfen962@aucklanduni.ac.nz}

\author{Paul Denny}
\orcid{0000-0002-5150-9806}
\affiliation{%
  \institution{University of Auckland}
  \streetaddress{Private Bag 92019}
  \city{Auckland}
  \country{New Zealand}
 }
\email{paul@cs.auckland.ac.nz}

\author{Burkhard C. W\"{u}nsche}
\orcid{0000-0002-8013-4118}
\affiliation{%
  \institution{University of Auckland}
  \streetaddress{Private Bag 92019}
  \city{Auckland}
  \country{New Zealand}
 }
\email{burkhard@cs.auckland.ac.nz}

\author{Andrew Luxton-Reilly}
\orcid{0000-0001-8269-2909}
\affiliation{%
  \institution{University of Auckland}
  \streetaddress{Private Bag 92019}
  \city{Auckland}
  \country{New Zealand}
}
\email{a.luxton-reilly@auckland.ac.nz}

\author{Jacqueline Whalley}
\orcid{0000-0001-7633-5200}
\affiliation{%
  \institution{Auckland University of Technology}
  \streetaddress{Private Bag 92006}
  \city{Auckland}
  \country{New Zealand}
}
\email{jacqueline.whalley@aut.ac.nz}

\setcopyright{rightsretained}
\copyrightyear{2024}
\acmYear{2024}
\acmConference{SA Educator's Forum '24}{December 03-06, 2024}{Tokyo, Japan}\acmBooktitle{SIGGRAPH Asia 2024 Educator's Forum (SA Educator's Forum '24), December 03-06, 2024}\acmDOI{10.1145/3680533.3697064}
\acmISBN{979-8-4007-1136-7/24/12}

\begin{document}

\begin{teaserfigure}
    \begin{tabular}{|p{0.32\textwidth}|p{0.64\textwidth}|}
        \hline
        \begin{tabular}{p{0.3\textwidth}}
            How many geometric objects are there in the following image? How many cubes? Spheres? Cylinders? \\
            \centering
            \includegraphics[width=0.25\textwidth]{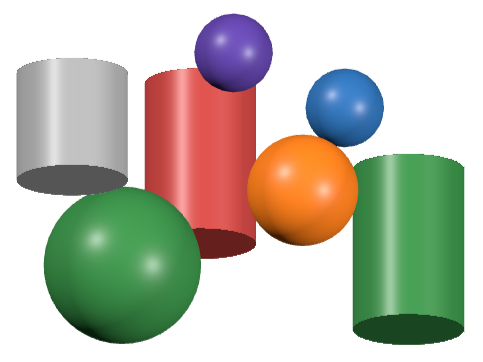}
        \end{tabular}
        &
        \begin{tabular}{p{0.62\textwidth}}
            The following two diagrams show several vectors in 3D space from two perspectives. Which of the colored vectors is coplanar (lying in the same plane) with both of the gray vectors? \\
            \centering
            \includegraphics[width=0.25\textwidth]{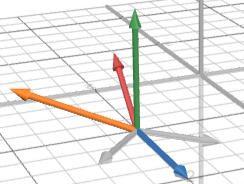}
            \includegraphics[width=0.25\textwidth]{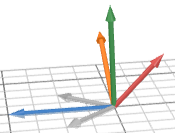}
        \end{tabular}
        \\
        \hline
    \end{tabular}
    \caption{Two questions from CG\_EASY that GPT-4o struggled with (2 out of 10 responses are correct for the left question, and 0 out of 10 responses are correct for the right question)}
    \label{fig:example-easy}
\end{teaserfigure}

\begin{abstract}


CG (Computer Graphics) is a popular field of CS (Computer Science), but many students find this topic difficult due to it requiring a large number of skills, such as mathematics, programming, geometric reasoning, and creativity. Over the past few years, researchers have investigated ways to harness the power of GenAI (Generative Artificial Intelligence) to improve teaching. In CS, much of the research has focused on introductory computing. A recent study evaluating the performance of an LLM (Large Language Model), GPT-4 (text-only), on CG questions, indicated poor performance and reliance on detailed descriptions of image content, which often required considerable insight from the user to return reasonable results. So far, no studies have investigated the abilities of LMMs (Large Multimodal Models), or multimodal LLMs, to solve CG questions and how these abilities can be used to improve teaching.


In this study, we construct two datasets of CG questions requiring varying degrees of visual perception skills and geometric reasoning skills, and evaluate the current state-of-the-art LMM, GPT-4o, on the two datasets. We find that although GPT-4o exhibits great potential in solving questions with visual information independently, major limitations still exist to the accuracy and quality of the generated results. We propose several novel approaches for CG educators to incorporate GenAI into CG teaching despite these limitations. We hope that our guidelines further encourage learning and engagement in CG classrooms.
\end{abstract}



\maketitle

\section{Introduction}
The advancement of GenAI (Generative Artificial Intelligence), especially LLMs (Large Language Models), has garnered global attention from the Computing Education research community \cite{denny2024computing}. LLMs excel at generating solutions that are typical of many programming-focused computing courses~\cite{finnie2022robots, denny2023conversing, finnie2023my, savelka2023thrilled}. However, since LLMs can only process textual inputs, they perform poorly in tasks requiring image inputs and/or visual and geometric processing skills~\cite{feng2024more}, which are essential in solving questions in CG (Computer Graphics)~\cite{suselo2017journey, rodrigues2021computer}.

LMMs (Large Multimodal Models), or VLMs (Visual Language Models), are extensions of LLMs that allow users to provide information in non-textual formats, such as images. An example is GPT-4o (GPT-4 Omni)~\cite{gpt4o}, an LMM developed by OpenAI that allows image inputs. The release of LMMs opened many new opportunities. With image inputs, users can provide visual context to the GenAI model, making human-AI interactions easier, and questions requiring visual context can now be asked effortlessly. This also possibly allows LMMs to solve questions requiring visual and geometric reasoning skills, such as those in CG.

Investigating the performance of LMMs, such as GPT-4o, on CG questions can provide insight into decisions and opportunities related to teaching CG. 
Past research suggests that the poor performance of GPT-4 in CG questions limits students' ability to misuse it~\cite{feng2024more}, but also makes it harder for CG instructors to use GenAI for teaching (e.g., by providing formative feedback, creating practice questions~\cite{feng2024can}, generating explanations). Evaluating the capabilities and limitations of GPT-4o in the context of CG enables educators to make more informed decisions about integrating GenAI into their teaching.



In this work, we investigate the visual perception and geometric reasoning capabilities of GPT-4o by using two datasets of CG-related questions. We compare the visual processing capabilities of GPT-4o to its textual processing capabilities and outline implications and recommendations for CG educators. Our study aims to answer the following Research Question:

\textit{How well can GPT-4o solve Computer Graphics questions requiring visual perception and geometric reasoning skills?}

\section{Related Work}
\subsection{LLMs in Education}
There has been substantial research on various GenAI models, with a prominent focus on LLMs. Past research showed impressive capabilities of LLMs in many subjects, such as reading comprehension~\cite{brown2020language}, law~\cite{katz2024gpt}, medicine~\cite{lievin2024can, nori2023capabilities}, and various other academic fields~\cite{ai4science2023impact}, thus bringing opportunities and challenges in many disciplines~\cite{abd2023large, tu2023should, yeadon2023impact}. In CS (Computer Science), LLMs achieved high performance in CS1~\cite{finnie2022robots, denny2023conversing}, CS2~\cite{finnie2023my}, and programming-related MCQs (multiple-choice questions)~\cite{savelka2023thrilled}, often surpassing average student performance. Additionally, LLMs are capable of assisting CS educators by generating educational material~\cite{leinonen2023using, liffiton2023codehelp, macneil2023experiences}, providing many new opportunities~\cite{bernstein2024nesting, denny2024prompt}. Despite the wide variety of tasks that LLMs can complete, there are still many areas in which they exhibit limited performance, such as reasoning~\cite{bang2023multitask}, visual programming~\cite{singla2023evaluating}, and Parsons Problems (a programming exercise where students reorder shuffled code blocks)~\cite{reeves2023evaluating}.

Understanding the capabilities of LLMs for CG is an area of active research. We evaluated the performance of GPT-4 (text-only) on assessment questions used in an undergraduate introductory CG course and found that GPT-4 produced correct solutions to only 42.1\% of the questions~\cite{feng2024more}. Another study assessed GPT-4's ability to generate code for a Ray Tracing application, and the results demonstrated a similar performance compared to the previous study (42\% accuracy)~\cite{feng2024can}. 


\subsection{Evaluations of LMMs}

We theorize that the low performance of LLMs for CG questions is due to CG requiring extensive visual-based reasoning skills, and LLMs struggle with these tasks due to their textual nature and lack of visual training data~\cite{singla2023evaluating, feng2024more}. LMMs allow users to provide visual context to the model directly. However, since LMMs are still relatively new, few studies have been conducted to measure their capabilities in various tasks.

Two early evaluation reports on GPT-4V (GPT-4 Vision~\cite{gpt4v}, the predecessor of GPT-4o) showcased its capabilities on queries requiring visual contexts in a wide variety of settings, such as visual math reasoning and code generation~\cite{wu2023early, yang2023dawn}. The results showed impressive visual-based reasoning skills of LMMs. Nevertheless, they often produce errors. Similar evaluation studies on more specialized areas showed that LMMs are somewhat capable of assisting in medical diagnoses~\cite{wu2023can}, map analysis~\cite{xu2024map}, and autonomous driving~\cite{driessen2024putting, wen2024road}. However, the consensus remains that there are significant limitations to the capabilities of LMMs, and more development is needed before they can reliably support real-world applications.

In the context of education, GPT-4V has been compared with its text-only counterpart, GPT-4 Turbo, on a specialized medical examination, and no statistically significant differences between the results were found between the two models~\cite{hirano2024gpt}, indicating that LMMs do not necessarily outperform LLMs. In a study more relevant to CS, the ability of GPT-4V to generate code based on UML diagrams was evaluated, and it was observed to perform well for simpler, single-class UML diagrams, but it failed to consistently generate correct code for more complex, multi-class UML diagrams~\cite{antal2024assessing}.

Despite the mediocre performance of LMMs on some educational tasks, the use of LMMs or similar applications can increase student performance and interest~\cite{zain2023use}. Effectively leveraging this in educational settings may lead to similar positive impacts.

\section{Methods}
\subsection{Overview}
We investigate the current capabilities of GPT-4o on CG questions by 1) collecting and creating CG questions; 2) converting them into the JSON format accepted by GPT-4o; 3) fetching responses from GPT-4o (through the OpenAI API) as attempts at answering the questions; 4) evaluating the correctness of the responses. We then interpret the results and make recommendations to CG educators about how LMMs can be used for CG teaching.

\subsection{Step 1: Collecting Questions}
Our first dataset derives from a previous study using GPT-4~\cite{feng2024more}. It contains 101 assessment questions used in a third-year introductory CG course, 68 of which are MCQs and 33 are programming questions. The questions are taken from the mid-semester tests and final exams of the 2022 and 2023 iterations of the course (i.e., four assessments in total). We refer to this dataset as CG\_TEST in this paper. The topics covered in the questions include but are not limited to introductory Linear Algebra, introductory OpenGL, Colors and Lighting, Illumination and Shading, Texture Mapping, Ray Tracing, 3D Modelling, Parametric Curves and Surfaces, and Image Processing.

Each assessment is split into Theory and Programming parts. Theory parts consist of MCQs of four or more options. Programming parts consist of programming questions that often require students to write code snippets, which are then executed against pre-written test cases. If all test cases are passed, then the student is awarded all marks allocated for the question. Otherwise, no marks are awarded. No partial marks are given.

Of all 101 questions, 67 contain no images, and 34 contain images. Although many of the questions contain no images, almost all questions require visual perception and geometric reasoning intelligence as the course focuses heavily on developing these skills and contains highly visual concepts. Several example questions are listed throughout the paper.

Since the assessment questions are quite technical and specialized, we also want to investigate GPT-4o's ability to process visual information without using specialized knowledge and whether this makes a difference in performance. Hence, we also created a small dataset containing 10 basic image-based CG-related short-answer questions, which we refer to as CG\_EASY in this paper. However, little to no CG background is required to answer these questions, and only common sense and a moderate amount of visual and geometric reasoning skills are needed. The questions involve identifying and counting geometric objects in a scene, light-surface interactions, basic 3D geometry, and basic 3D transformations (translations and rotations). Figure \ref{fig:example-easy} shows two example questions from this dataset.

The two datasets used in this study are publicly available through the link provided in Section \ref{link}.

\subsection{Step 2: Converting to JSON}
GPT-4o allows for inputs in various formats, such as images from publicly accessible URLs, and combinations of multimodal content as single inputs, such as interweaving texts and images. Multimodal inputs follow the JSON format shown in Figure \ref{fig:template}.

\begin{figure}[t]
    \centering
    \begin{tabular}{|p{0.45\textwidth}|}
    \hline
    [ \\
    \quad   \{ "type": "text", "text": "[TEXT]" \}, \\
    \quad   \{ "type": "image\_url", "image\_url": \{ "url": "[IMAGE\_URL]", \} \}, \\
    \quad   ... \\
    ] \\
    \hline
    \end{tabular}
    \caption{GPT-4o multimodal input JSON format}
    \label{fig:template}
\end{figure}

The questions collected from CG assessments contain texts, mathematical formulas, and images, but they are not in the format accepted for multimodal inputs. Hence, some preprocessing needs to occur before the questions can be processed by GPT-4o.

Texts can be copied directly into the \textit{``text''} field corresponding to \textit{``type'': ``text''}. Mathematical formulas are replaced by their corresponding LaTeX commands. An alternative method would be to replace formulas with their textual counterparts, for example, replacing ``$2\times3$'' (written with LaTeX commands) with text symbols such as ``2x3'' and ``2*3'', but such replacements can be inconsistent. 
Hence, TeX commands are used to keep conversions consistent. Figure \ref{fig:example-formulas} shows an example of this conversion.

\begin{figure}
    \centering
    \begin{tabular}{|p{0.45\textwidth}|}
        \hline
        \begin{tabular}{p{0.43\textwidth}}
            Given is a plane 3x+2y-z=3 and a ray
            $$
            p(t) = 
            \begin{pmatrix}
            1 \\ 0 \\ 1
            \end{pmatrix}
            + t * 
            \begin{pmatrix}
            -1 \\ c \\ 0
            \end{pmatrix}
            $$
            For what value of c is the ray parallel to the plane? \\
            Select one: \\
            a. c=0; b. c=1.5; c. c=0.5; d. c=1; e. c=-0.5
        \end{tabular}
        \\
        \hline
        \begin{tabular}{p{0.43\textwidth}}
            [ \\
            \quad   \{ \\
            \quad   \quad   "type": "text", "text": """ \\
            Given is a plane 3x+2y-z=3 and a ray \\ 
            \$\$p(t)=\textbackslash begin\{pmatrix\}1\textbackslash\textbackslash0\textbackslash\textbackslash1\textbackslash end\{pmatrix\}+t*\textbackslash begin\{pmatrix\} \\
            -1\textbackslash\textbackslash c\textbackslash\textbackslash0\textbackslash end\{pmatrix\}.\$\$ \\
            For what value of c is the ray parallel to the plane? \\
            Select one: \\
            a. c=0; b. c=1.5; c. c=0.5; d. c=1; e. c=-0.5 \\
            \quad   \quad   """ \\
            \quad   \}, \\
            ] \\
        \end{tabular}
        \\
        \hline
    \end{tabular}
    \caption{An example assessment question containing formulas (top); the textual version of the question using LaTeX commands, in the GPT-4o text-only input format (bottom)}
    \label{fig:example-formulas}
\end{figure}

\begin{figure}
    \centering
    \begin{tabular}{|p{0.2\textwidth}|p{0.19\textwidth}|}
        \hline
        \multicolumn{1}{|c|}{\includegraphics[width=0.2\textwidth]{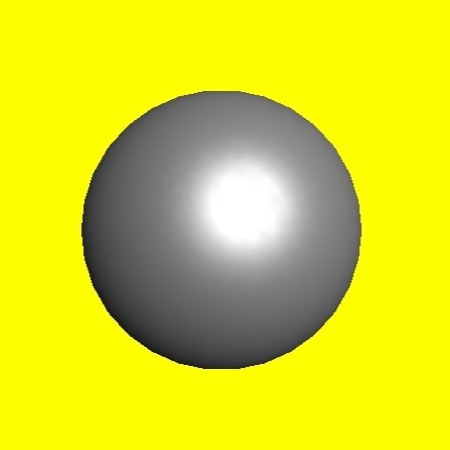}} & \multicolumn{1}{|c|}{\includegraphics[width=0.2\textwidth]{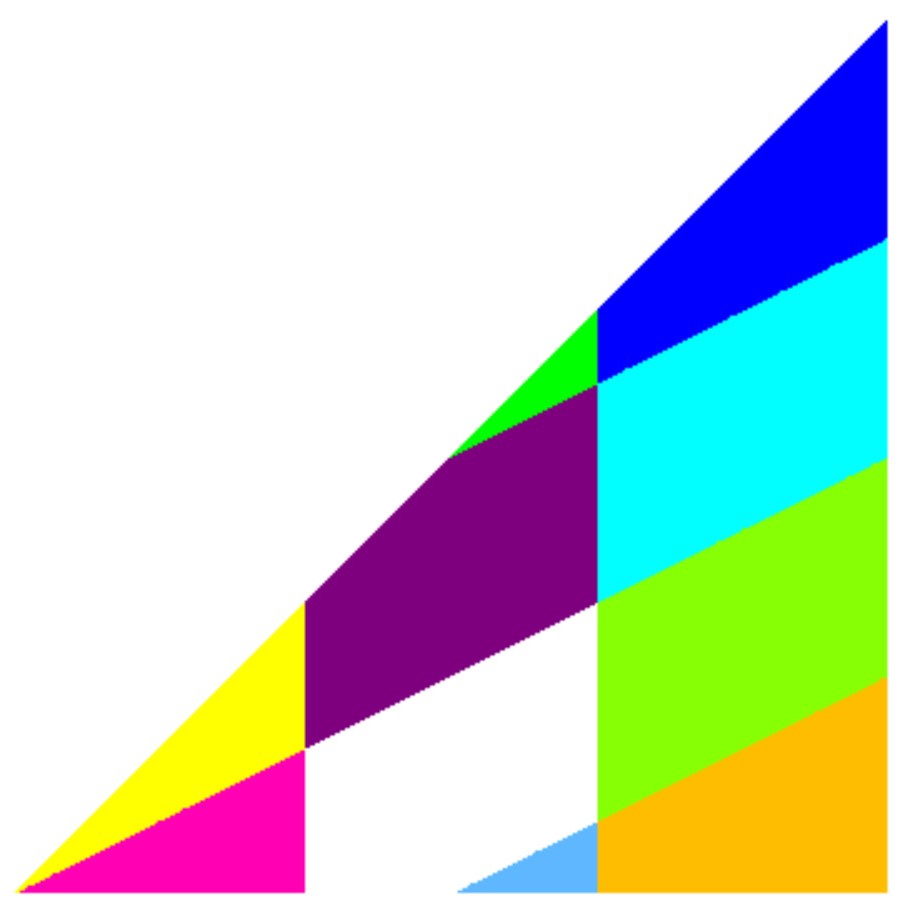}} \\
        The background is bright yellow, and a gray sphere is shown with a medium level of lighting, with a bright white highlight covering a small portion of the surface of the sphere. & The image shows a shifted version of the base image. Using grid coordinates from the base image, where the bottom left corner of the base image is (0, 0) and the top right corner of the base image is (4, 4), then the bottom left corner is (0, 2), the bottom right corner is (3, 0), and the top right corner is (3, 4). \\
        \hline
    \end{tabular}
    \caption{Two example images used in CG\_TEST and their corresponding textual descriptions}
    \label{fig:example-descriptions}
\end{figure}

\begin{figure*}
    \centering
    \begin{tabular}{|p{0.33\textwidth}|p{0.24\textwidth}|p{0.37\textwidth}|}
        \hline
        \begin{tabular}{p{0.31\textwidth}}
            Given a function drawShape() which draws a wireframe representation of the letter "L" in the xy-plane as shown in the image below. \\
            \includegraphics[width=0.25\textwidth]{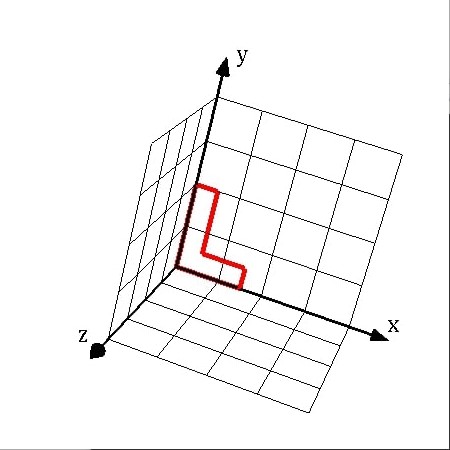} \\
            Please write OpenGL code to transform this shape such that you obtain the scene displayed in the image below: \\
            \includegraphics[width=0.25\textwidth]{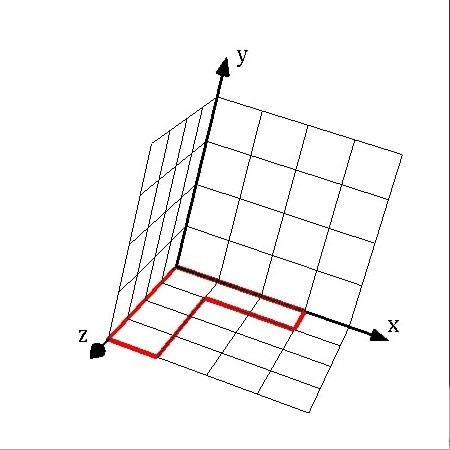} \\
            IMPORTANT: \\
            Please only use OpenGL transformations, e.g. glScalef, glTranslatef, glRotatef. \\
            Please do NOT draw the shape itself - this is done automatically by the uploaded code. \\
        \end{tabular}
        &
        \begin{tabular}{p{0.22\textwidth}}
            [ \\
            \quad   \{ \\
            \quad   \quad   "type": "text", "text": """ \\
            Given is a function drawShape() which draws a wireframe representation of the letter "L" in the xy-plane as shown in the image below. \\
            \\
            Image description: A letter L is placed on the x-y plane. The vertices on the shape are p1 = (0, 0, 0), p2 = (0, 2, 0), p3 = (1.5, 0, 0). \\
            \\
            Please write OpenGL code to transform this shape such that you obtain the scene displayed in the image below: \\
            \\
            Image description: A letter L with twice the size is placed on the x-z plane. The vertices on the shape are p1 = (0, 0, 0), p2 = (0, 0, 4), p3 = (3, 0, 0). \\
            \\
            IMPORTANT: \\
            Please only use OpenGL transformations, e.g. glScalef, glTranslatef, glRotatef. \\
            Please do NOT draw the shape itself - this is done automatically by the uploaded code. \\
            \quad   \quad   """ \\
            \quad   \}, \\
            ] \\
        \end{tabular}
        &
        \begin{tabular}{p{0.35\textwidth}}
            [ \\
            \quad   \{ \\
            \quad   \quad   "type": "text", "text": """ \\
            Given is a function drawShape() which draws a wireframe representation of the letter "L" in the xy-plane as shown in the image below. \\
            \quad   \quad   """ \\
            \quad   \}, \{ \\
            \quad   \quad   "type": "image\_url", "image\_url": \{ "url": """ \\
            \url{https://raw.githubusercontent.com/TFPlusPlus/GPT-4V-vs.-CG/main/CG_Assessments_Images/2022b15-1.jpg} \\
            \quad   \quad   \quad   """ \\
            \quad   \quad   \} \\
            \quad   \}, \{ \\
            \quad   \quad   "type": "text", "text": """ \\
            Please write OpenGL code to transform this shape such that you obtain the scene displayed in the image below: \\
            \quad   \quad   """ \\
            \quad   \}, \{ \\
            \quad   \quad   "type": "image\_url", "image\_url": \{ "url": """ \\
            \url{https://raw.githubusercontent.com/TFPlusPlus/GPT-4V-vs.-CG/main/CG_Assessments_Images/2022b15-2.jpg} \\
            \quad   \quad   \quad   """ \\
            \quad   \quad   \} \\
            \quad   \}, \{ \\
            \quad   \quad   "type": "text", "text": """ \\
            IMPORTANT: \\
            Please only use OpenGL transformations, e.g. glScalef, glTranslatef, glRotatef. \\
            Please do NOT draw the shape itself - this is done automatically by the uploaded code. \\
            \quad   \quad   """ \\
            \quad   \} \\
            ] \\
        \end{tabular}
        \\
        \hline
    \end{tabular}
    \caption{An example assessment question containing images (left); the textual version of the question using image descriptions, in the GPT-4o text-only input JSON format (middle); the multimodal version of the question using the real images, in the GPT-4o multimodal input JSON format (right)}
    \label{fig:example-images}
\end{figure*}

Although we can directly feed images to GPT-4o, we also want to investigate any differences in performance on image-based questions with its text-only capabilities. Past literature has used image descriptions to encompass all relevant information in the images, which is then fed as input to LLMs in place of the images~\cite{yang2022empirical}. Our previous study that evaluated CG questions also used this strategy before the publication of LMMs~\cite{feng2024more}. Therefore, for every image in our dataset of assessment questions, we can replace it with a textual description of the image, at the level of detail a capable student could produce, containing all the information necessary to solve the corresponding question.
Two examples are shown in Figure \ref{fig:example-descriptions}. Then, for every question containing images, we constructed two JSON objects in their corresponding 
input formats for GPT-4o: text-only using textual descriptions and multimodal using real images. An example is shown in Figure \ref{fig:example-images}.

\subsection{Step 3: Fetching Responses}
After the questions are converted to JSON objects, the data is sent to GPT-4o via the OpenAI API, to which the model responds with its answers. Each JSON object is sent 10 times, and 10 responses are received, which are treated as 10 independent attempts. For every question containing images, two separate JSON objects are sent (text-only version and multimodal version), and 20 responses are received for that question, 10 for each version. The model's temperature is set to 0.75, which is reported to perform well on previous, similar studies~\cite{feng2024more, pursnani2023performance}. The system message we use in this study is ``You are a helpful assistant, and you are knowledgeable in Computer Graphics. When you answer a multiple-choice question, you state your selected option explicitly while providing a concise and accurate explanation.''.

\subsection{Step 4: Evaluating Correctness}
The responses from the GPT-4o are then evaluated for correctness. No partial marks are given (responses are categorized as correct or incorrect).

For MCQs, responses are marked as correct if they state the correct option or the letter associated with the correct option. The responses usually contain explanations of their solutions, but they are not required to be considered correct.

For programming questions, responses are marked as correct if they contain the correct solution code that can be copied and pasted into the AAT (Automated Assessment Tool) used in the assessments and pass all test cases~\cite{wunsche2018automatic,wunsche2019automatic}. We allow for some deletions from the generated code solutions, such as boilerplate code which is often present in outputs, as boilerplate code is already supplied by the AAT.

Furthermore, the accuracy is evaluated in two ways for programming questions: ''1 attempt'' and ``10 attempts''. Under the conditions of these assessments, students are allowed to submit their code solutions for programming questions as many times as they want with no penalties, and they obtain full marks for the question as long as one of their attempts passes all test cases~\cite{wunsche2018automatic}. The ``10 attempts'' marking scheme is used to mimic this setting. Alternatively, ``1 attempt'' mimics the setting where the student has only 1 attempt for each question, and it measures the expected score from 10 separate attempts. This is done by taking the average of the 10 responses for every question.

\begin{table}[t]
    \centering
    \caption{Percentages of correct responses from GPT-4 (text-only)~\cite{feng2024more} and GPT-4o for questions in various categories (the higher percentage is marked in bold)}
    \begin{tabular}{|p{0.28\textwidth}|r|r|}
        \hline
        Category & GPT-4 & GPT-4o \\
        \hline
        \hline
        CG\_TEST: All questions & 42.1\% & \textbf{50.1\%} \\
        \hline
        \hline
        CG\_TEST: MCQs & 53.5\% & \textbf{62.6\%} \\
        \hline
        CG\_TEST: Programming (1 attempt) & 27.1\% & \textbf{31.8\%} \\
        \hline
        CG\_TEST: Programming (10 attempts) & \textbf{53.6\%} & 50.9\% \\
        \hline
        \hline
        CG\_TEST: No images & 60.0\% & \textbf{67.9\%} \\
        \hline
        CG\_TEST: Images (textual descriptions) & \textbf{36.5\%} & 35.6\% \\
        \hline
        CG\_TEST: Images (real images) & N/A & \textbf{29.4\%} \\
        \hline
        \hline
        CG\_EASY: All questions & N/A & \textbf{62.0\%} \\
        \hline
    \end{tabular}
    \label{tab:percentages}
\end{table}

\section{Results}
For CG\_TEST, out of all 1350 responses received from GPT-4o (67 text-only questions, 34 image-based questions converted to text-only using textual descriptions, 34 image-based questions using multimodal input, 10 responses each), 676 responses are marked as correct, which is 50.1\% of all responses. Out of 800 responses to MCQs (56 text-only MCQs, two versions of 12 image-based questions), 501 responses are correct (62.6\%). Of 550 responses to programming questions (11 text-only questions, two versions of 22 image-based questions), 175 responses are correct (31.8\%). There are 55 groups of 10 responses to programming questions, each corresponding to one programming question, and 28 out of the 55 groups contained at least one correct response (50.9\%). Of 670 responses to text-only questions (67 text-only questions), 455 are correct (67.9\%). Of 340 responses to image-based questions using textual descriptions (34 image-based questions using textual descriptions), 121 are correct (35.6\%). Of 340 responses to image-based questions using real images (34 image-based questions using real images), 100 are correct (29.4\%). For CG\_EASY, out of 100 responses from GPT-4o (10 image-based questions using real images), 62 are correct (62.0\%). These results are summarized in Table \ref{tab:percentages}.

\section{Discussion}
\subsection{Overall Results}
Overall, GPT-4o answers around half of the queries from CG assessments correctly, i.e., it is not a reliable source of answers for CG assessments or specialized CG questions.
A slightly higher accuracy is achieved for questions in the CG\_EASY dataset. However, it is still well below human performance, as these questions can be solved easily with common sense and minimal visual and geometric reasoning skills. 
This indicates a lack of consistency and reliability in GPT-4o's visual and geometric reasoning skills. Users should be mindful of these risks for questions requiring these skills.

The performance of GPT-4o on MCQs is higher than that on programming questions. This may be because MCQs are generally easier than programming questions, as they usually only involve one or two specific concepts. In contrast, programming questions are more complex and require more steps and critical thinking. 
Additionally, even if GPT-4o does not ``know'' the correct answer, it can still return a correct answer due to chance (25\% chance for MCQs with 4 options, 20\% for MCQs with 5 options).

GPT-4o performs much better on questions containing no images than those containing images. This indicates that despite possessing visual processing power, GPT-4o is still vastly superior at textual processing than visual processing, even if the question context requires visual and geometric reasoning skills. The question difficulty may also contribute to the difference in performance, as more difficult questions often require images to illustrate the context, so questions containing images may, on average, be more difficult than those containing none.

Comparing the results for image-based questions, queries using textual descriptions outperform those using real images. This suggests that prompts written by humans can distinguish important features in images that GPT-4o fails to do, and descriptions of these features may help improve the model's performance. Another interpretation is that through describing the images, the human, instead of the model, does some of the visual processing, so the visual processing required for GPT-4o is reduced. Since GPT-4o is observed to underperform in visual processing tasks, this reduces the risk of GPT-4o making a mistake. 

A similar past study evaluated the performance of GPT-4 (text-only) on the CG\_TEST dataset~\cite{feng2024more}. Comparing the results, we see an increase in the performance of ``All questions'', ``MCQs'', and ``Contains no images''. This indicates that GPT-4o has, on average, improved its capability to answer CG questions compared to its predecessor GPT-4, especially on textual questions. ``Programming (1 attempt)'' shows a slight increase in performance, but ``Programming (10 attempts)'' does not reach past performance. This suggests that GPT-4o is more consistently correct on questions that it is confident in, but the difference may simply be due to chance and is too small to be conclusive evidence. The performance of image-based questions using textual descriptions also does not improve, indicating a lack of improvement in visual reasoning skills from textual descriptions.

\subsection{Common Characteristics of Responses}
Please note that the behavior of the responses is dependent on the system message used with the query, and queries using different system messages may not show the following characteristics.

\begin{sloppypar}
\subsubsection{Varying lengths of explanations}
The responses to the CG\_TEST dataset are generally lengthy and detailed, with most explanations to questions reaching more than 10 lines of text and some even reaching 30 lines. Additionally, GPT-4o would often make mistakes but continue to elaborate along incorrect lines of thinking, which can confuse students and reduce learning. Common errors made by GPT-4o for this dataset are conceptual errors (e.g., using incorrect concepts, hallucinating false facts), mathematical errors (e.g., incorrectly substituting values into formulas, incorrectly expanding expressions, calculation errors), and logical errors (e.g., stating fallacious causal relationships).
\end{sloppypar}

The responses to CG\_EASY are much shorter, typically only around 1-10 lines long, since solutions are often straightforward and do not require complex explanations (although the responses can still be incorrect).

\begin{figure}
    \centering
    \includegraphics[width=0.19\textwidth]{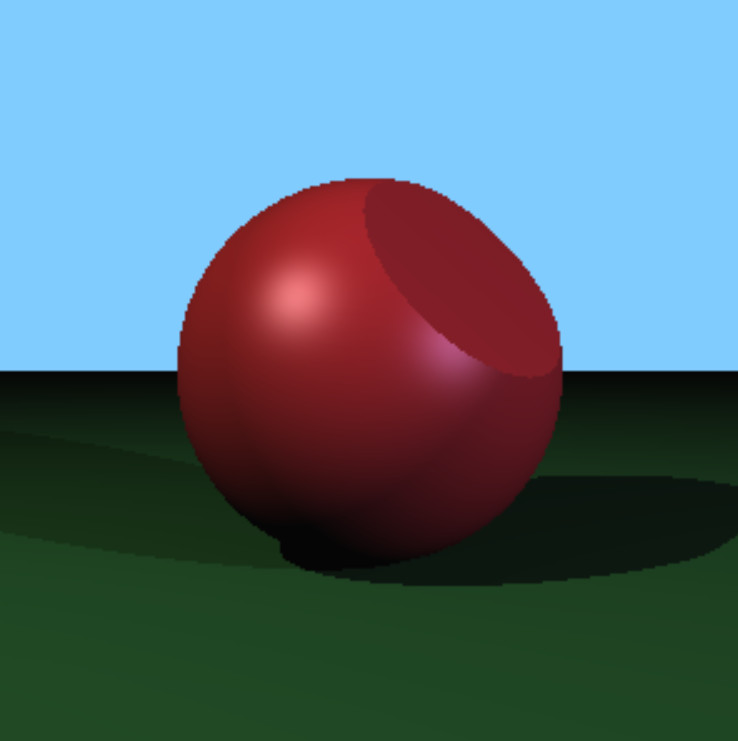}
    \caption{An image used in a programming question asking students to write code for intersecting a ray with a sphere cut by a plane. The function takes the ray's start point and direction as input and returns the intersection point.}
    \label{fig:cut-sphere}
\end{figure}

\begin{figure}
    \centering
    \includegraphics[width=0.19\textwidth]{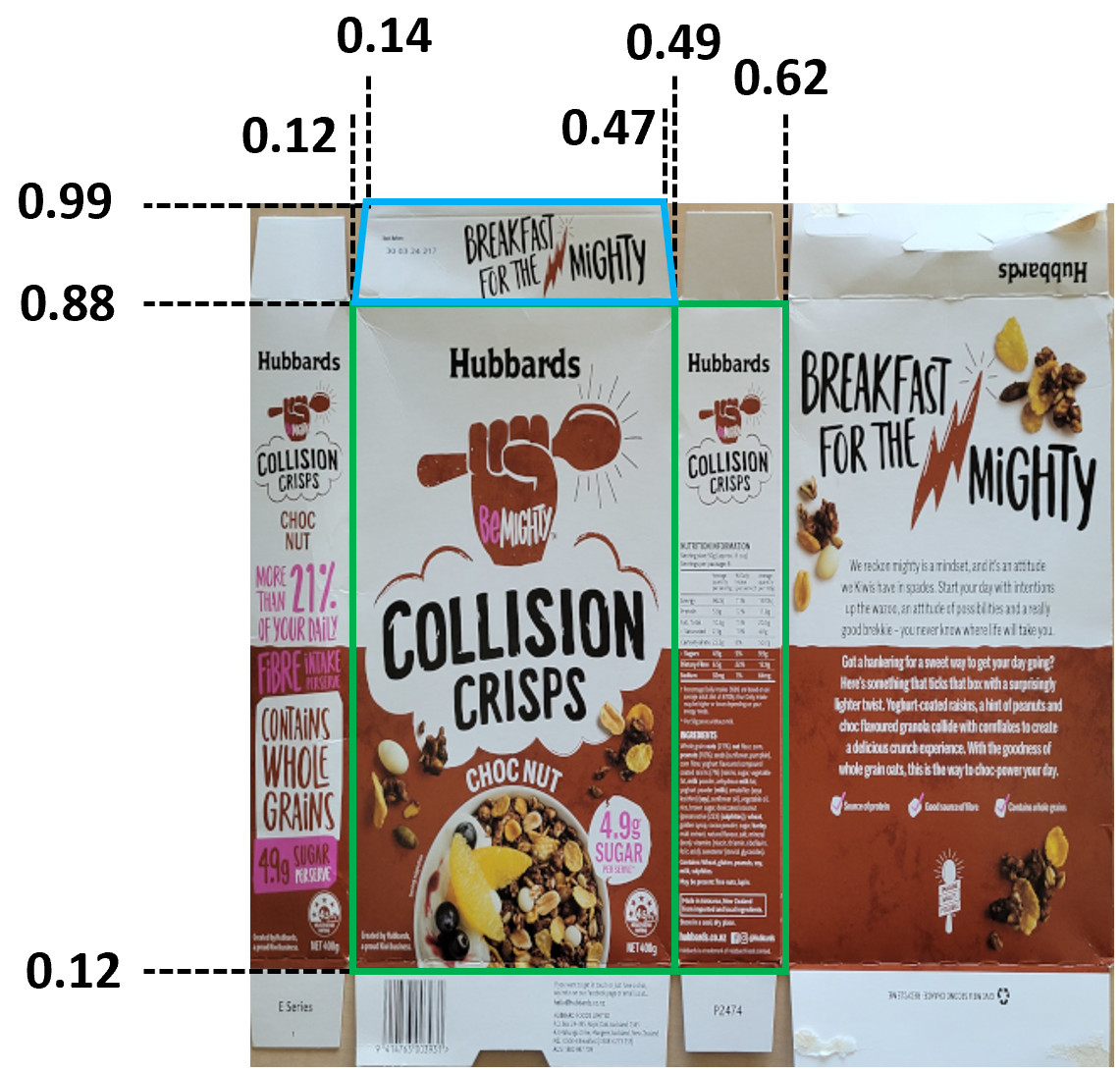}
    \includegraphics[width=0.19\textwidth]{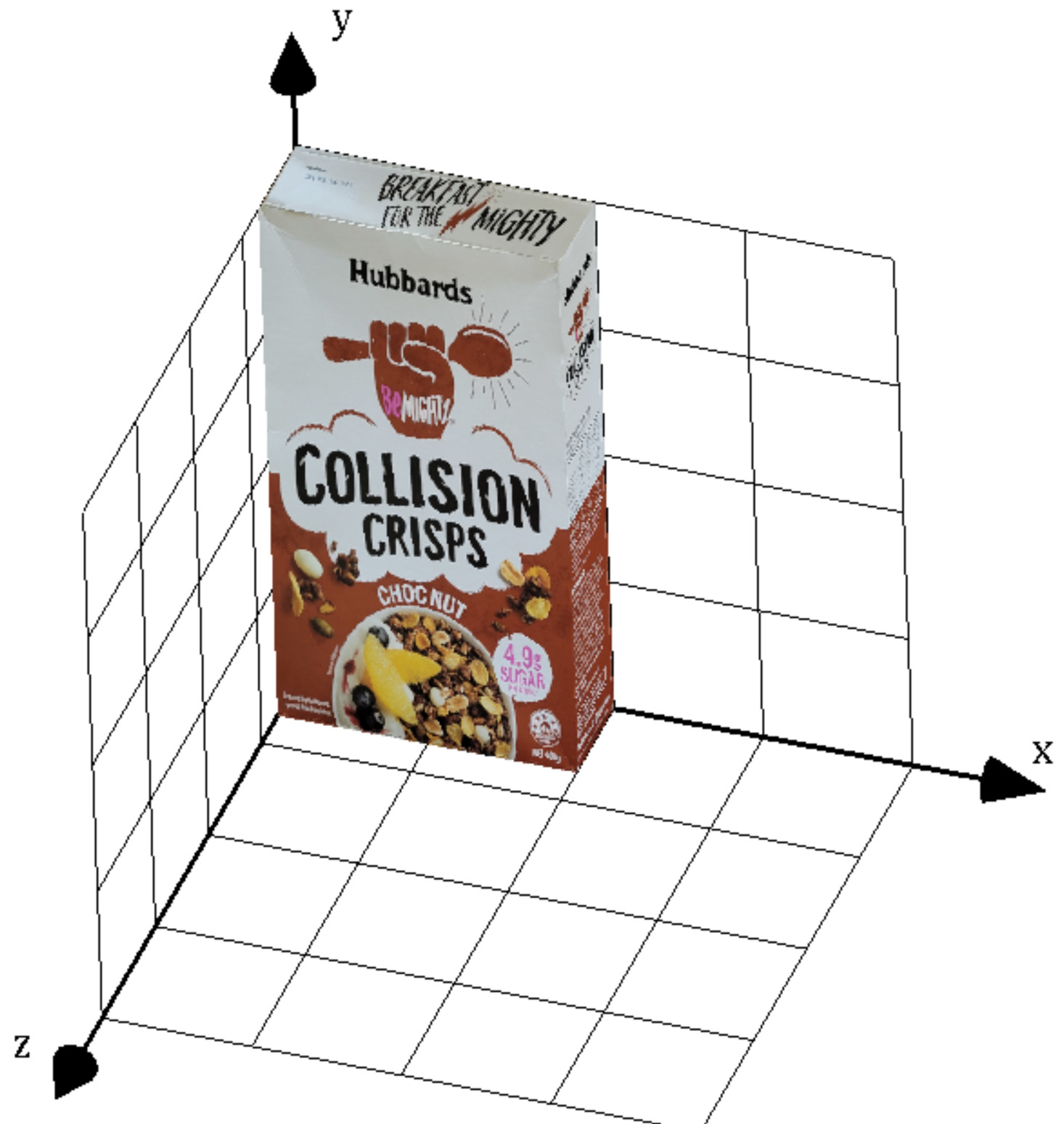}
    \caption{Two images used in a programming question asking students to map a texture image (left) onto a polygon mesh (right).}
    \label{fig:texture}
\end{figure}

\begin{sloppypar}
\subsubsection{Unnecessary code}
For programming questions in the CG\_TEST dataset, the generated code snippets can be unnecessarily long and contain large amounts of boilerplate code. For example, the solution to a 3D programming question (shown in Figure \ref{fig:transformation}) is simply 3 lines of OpenGL code, calling the functions glRotatef(), glScalef(), glTranslatef() once each. However, most generated solutions for this question exceed 20 lines of code and include boilerplate code supplied by the AAT, such as a main() function. There are also many instances where the questions state that some classes and functions are already provided, but GPT-4o still includes those classes and functions with their full implementations, which leads to redefinition errors when the unmodified code solutions are executed, and these extra code snippets have to be manually removed when testing for correctness.
\end{sloppypar}

\subsubsection{Failure to follow specific instructions}
In addition to disregarding statements specifying the existing classes and functions, GPT-4o often fails to follow the correct syntax to call existing functions. For example, a common function supplied by the debugger is ``dot(Vector v1, Vector v2)'' for calculating the dot product of two vectors. The syntax for calling this function is always specified in the questions, but GPT-4o often fails to follow the syntax and instead writes ``v1.dot(v2)''. More than 40 responses out of the total 550 for programming questions include errors of this kind.

\subsection{Specific Observations}
\subsubsection{Breakthroughs in difficult questions}
The CG\_TEST dataset contains several questions GPT-4 could not answer correctly in previous studies~\cite{feng2024can, feng2024more}. One example is a programming question for ray tracing a cut sphere (shown in Figure \ref{fig:cut-sphere}), which GPT-4 answered incorrectly for all 30 attempts across two studies. In this study, GPT-4o answered this question correctly (with complete working code) in 1 out of the 10 responses for the textual description version and also 1 out of 10 for the image version. This is impressive since fewer than 5\% of students could answer this question with unlimited attempts in an exam.

The CG\_TEST dataset also contains a texture mapping programming question (shown in Figure \ref{fig:texture}), GPT-4 (text-only) could consistently solve this question, but this could be due to the textual descriptions of the images, and the visual processing that the human did by extracting the coordinates of the faces when describing the images. However, without the help of textual descriptions, GPT-4o can successfully extract the coordinates of the faces from the images and solve this question in 1 out of the 10 responses.

We acknowledge that the breakthrough performance on these two questions could potentially be only due to chance, but when combined with all other results, we suggest that over the past year, GPT-4 has improved (multimodal vs. text-only) in performance for CG questions.

\subsubsection{Challenges in answering questions in CG\_EASY}
GPT-4o can successfully solve 62.0\% of the image-based questions in CG\_EASY without any human assistance or textual descriptions for the images, which is certainly an impressive performance. However, there are still some questions that GPT-4o struggles with, such as the two questions shown in Figure \ref{fig:example-easy}.

For the question shown on the left of Figure \ref{fig:example-easy}, GPT-4o answers correctly in only 2 out of the 10 responses. In the 8 other attempts, GPT-4o states that there are either 8 or 9 objects in the image, and in most cases, it identifies 4 cylinders. However, when directly asking about the number of cylinders in the image, GPT-4o answers correctly 9 out of 10 times. We theorize that the added complexity of the question may have confused GPT-4o, and this reduction in performance may not be related to its visual perception skills.

For the question on the right of Figure \ref{fig:example-easy}, GPT-4o incorrectly answers ``orange'' in all 10 attempts. Although it sometimes states that the blue vector is also coplanar with the gray vectors, it always perceives the orange vector as coplanar, hence they cannot be marked as correct. From the results of this question, we suggest that although GPT-4o has modest visual perception skills, it still lacks geometric reasoning skills.

\subsubsection{Challenges in answering 3D transformation questions}
\begin{figure}
    \centering
    \includegraphics[width=0.19\textwidth]{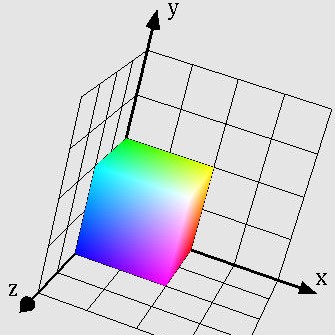}
    \includegraphics[width=0.19\textwidth]{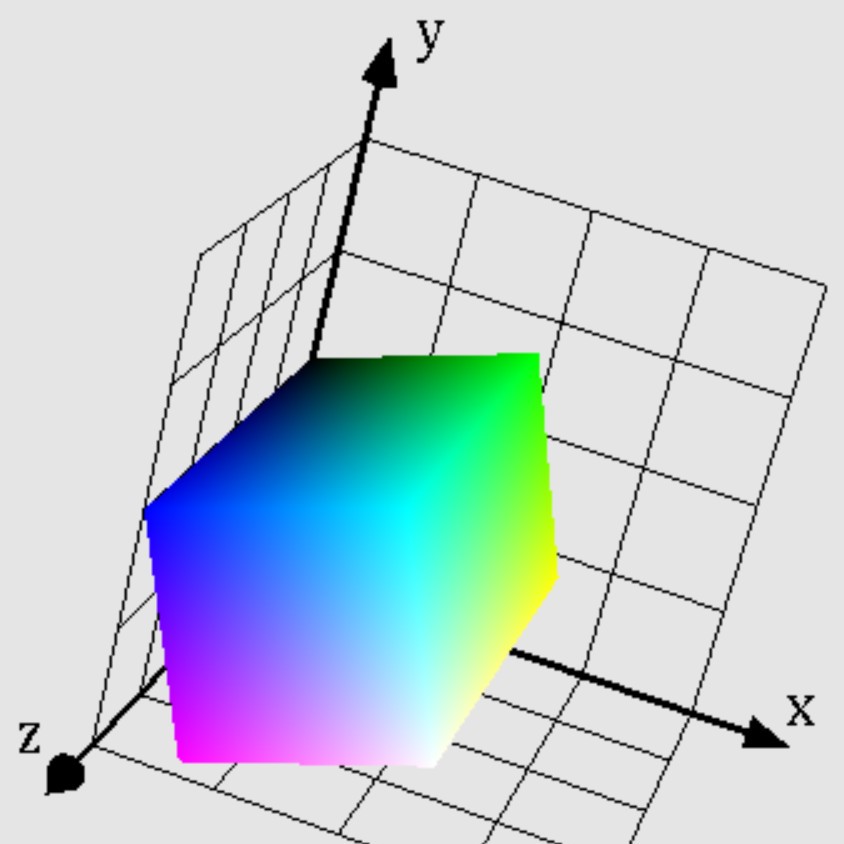}
    \caption{Two images used in a programming question asking students to transform a 3D color cube from its original position (left) to a new position (right) using OpenGL functions glScalef, glTranslatef, glRotatef. A textual description of the 3D transformation is also provided in the question.}
    \label{fig:transformation}
\end{figure}

A question type that GPT-4 and GPT-4o struggle with is programming questions related to 3D transformations, one of which is shown in Figure \ref{fig:transformation}. All 40 attempts from this study and the previous study provide incorrect code solutions for this question, despite the correct solution only being 3 lines of code. This is further evidence that GPT-4 and GPT-4o lack geometric reasoning skills, which are essential in solving this question.

\subsection{Implications}
\subsubsection{GenAI models are unreliable in visual question-answering}
The results of our study suggest that GPT-4o, or LMMs in general, may not reliably answer CG questions requiring visual perception skills and especially geometric reasoning skills. However, this does not mean that GenAI models cannot be used to improve learning. For example, CG educators can write or generate image descriptions for CG problems and ask students to evaluate the quality of the descriptions and/or improve the descriptions to enable GenAI models to solve the original problems. Additionally, LMMs are also useful for improving self-reflective practice~\cite{kumar2024supporting}.

Conversely, CG educators should also raise student awareness of the limitations of GenAI for CG questions and the importance of critically evaluating the generated solutions. For CG educators who are opposed to the use of GenAI for teaching and learning purposes, since GPT-4o performs more poorly on image-based questions than textual questions, greater use of image-based questions may discourage students from using GenAI and encourage independent thinking and learning.

\subsubsection{A new exercise: Spot the error}
An exercise for CG educators is to use incorrect AI-generated solutions to CG questions and ask students to find the errors in these solutions. This can simultaneously encourage students to reflect critically on their understanding of the topics and also raise awareness of the limitations of GenAI.


\subsubsection{Prompt engineering for more accurate and helpful responses}
In this study, we directly used the question texts (after formatting) as prompts for the GenAI model. Recent research suggests that GenAI can achieve higher performance through better prompting strategies, such as splitting each question into smaller subquestions or asking GenAI to explain step by step~\cite{kojima2022large, denny2023conversing}. Different system messages can also be used to achieve different performances and characteristics. This could also be a good learning task for students, i.e., develop prompting strategies to solve complex CG questions.

\section{Conclusion}
In this study, we constructed two datasets of CG assessment and basic visual CG-related questions requiring varying degrees of visual perception skills and geometric reasoning skills.  We evaluated the performance of GPT-4o on these two datasets. Although GPT-4o has improved in performance on visual questions compared to predecessor models, it still lacks the visual processing power to provide reliable academic support to CG students and, in general, real-world applications requiring visual understanding. We also described several common characteristics exhibited by GPT-4o in its responses and outlined various specific questions on which GPT-4o performed well or poorly. Finally, we suggested some implications for CG education and provided recommendations to CG educators on utilizing LMMs to improve CG teaching.


\section{Resources}
\label{link}
All images, textual descriptions, and JSON objects can be accessed through this link: \url{https://github.com/TFPlusPlus/GPT-4V-vs.-CG}.

\balance
\bibliographystyle{ACM-Reference-Format}
\bibliography{ref}


\begin{thebibliography}{42}


\ifx \showCODEN    \undefined \def \showCODEN     #1{\unskip}     \fi
\ifx \showDOI      \undefined \def \showDOI       #1{#1}\fi
\ifx \showISBNx    \undefined \def \showISBNx     #1{\unskip}     \fi
\ifx \showISBNxiii \undefined \def \showISBNxiii  #1{\unskip}     \fi
\ifx \showISSN     \undefined \def \showISSN      #1{\unskip}     \fi
\ifx \showLCCN     \undefined \def \showLCCN      #1{\unskip}     \fi
\ifx \shownote     \undefined \def \shownote      #1{#1}          \fi
\ifx \showarticletitle \undefined \def \showarticletitle #1{#1}   \fi
\ifx \showURL      \undefined \def \showURL       {\relax}        \fi
\providecommand\bibfield[2]{#2}
\providecommand\bibinfo[2]{#2}
\providecommand\natexlab[1]{#1}
\providecommand\showeprint[2][]{arXiv:#2}

\bibitem[Abd-Alrazaq et~al\mbox{.}(2023)]%
        {abd2023large}
\bibfield{author}{\bibinfo{person}{Alaa Abd-Alrazaq}, \bibinfo{person}{Rawan AlSaad}, \bibinfo{person}{Dari Alhuwail}, \bibinfo{person}{Arfan Ahmed}, \bibinfo{person}{Padraig~Mark Healy}, \bibinfo{person}{Syed Latifi}, \bibinfo{person}{Sarah Aziz}, \bibinfo{person}{Rafat Damseh}, \bibinfo{person}{Sadam~Alabed Alrazak}, \bibinfo{person}{Javaid Sheikh}, {et~al\mbox{.}}} \bibinfo{year}{2023}\natexlab{}.
\newblock \showarticletitle{Large Language Models in Medical Education: Opportunities, Challenges, and Future Directions}.
\newblock \bibinfo{journal}{\emph{JMIR Medical Education}} \bibinfo{volume}{9}, \bibinfo{number}{1} (\bibinfo{year}{2023}), \bibinfo{pages}{e48291}.
\newblock


\bibitem[AI4Science and Quantum(2023)]%
        {ai4science2023impact}
\bibfield{author}{\bibinfo{person}{Microsoft~Research AI4Science} {and} \bibinfo{person}{Microsoft~Azure Quantum}.} \bibinfo{year}{2023}\natexlab{}.
\newblock \showarticletitle{The impact of large language models on scientific discovery: a preliminary study using gpt-4}.
\newblock \bibinfo{journal}{\emph{arXiv preprint arXiv:2311.07361}} (\bibinfo{year}{2023}).
\newblock


\bibitem[Antal et~al\mbox{.}(2024)]%
        {antal2024assessing}
\bibfield{author}{\bibinfo{person}{G{\'a}bor Antal}, \bibinfo{person}{Rich{\'a}rd Voz{\'a}r}, {and} \bibinfo{person}{Rudolf Ferenc}.} \bibinfo{year}{2024}\natexlab{}.
\newblock \showarticletitle{Assessing GPT-4-Vision's Capabilities in UML-Based Code Generation}.
\newblock \bibinfo{journal}{\emph{arXiv preprint arXiv:2404.14370}} (\bibinfo{year}{2024}).
\newblock


\bibitem[Bang et~al\mbox{.}(2023)]%
        {bang2023multitask}
\bibfield{author}{\bibinfo{person}{Yejin Bang}, \bibinfo{person}{Samuel Cahyawijaya}, \bibinfo{person}{Nayeon Lee}, \bibinfo{person}{Wenliang Dai}, \bibinfo{person}{Dan Su}, \bibinfo{person}{Bryan Wilie}, \bibinfo{person}{Holy Lovenia}, \bibinfo{person}{Ziwei Ji}, \bibinfo{person}{Tiezheng Yu}, \bibinfo{person}{Willy Chung}, {et~al\mbox{.}}} \bibinfo{year}{2023}\natexlab{}.
\newblock \showarticletitle{A multitask, multilingual, multimodal evaluation of chatgpt on reasoning, hallucination, and interactivity}.
\newblock \bibinfo{journal}{\emph{arXiv preprint arXiv:2302.04023}} (\bibinfo{year}{2023}).
\newblock


\bibitem[Bernstein et~al\mbox{.}(2024)]%
        {bernstein2024nesting}
\bibfield{author}{\bibinfo{person}{Seth Bernstein}, \bibinfo{person}{Paul Denny}, \bibinfo{person}{Juho Leinonen}, \bibinfo{person}{Lauren Kan}, \bibinfo{person}{Arto Hellas}, \bibinfo{person}{Matt Littlefield}, \bibinfo{person}{Sami Sarsa}, {and} \bibinfo{person}{Stephen Macneil}.} \bibinfo{year}{2024}\natexlab{}.
\newblock \showarticletitle{"Like a Nesting Doll": Analyzing Recursion Analogies Generated by CS Students Using Large Language Models}. In \bibinfo{booktitle}{\emph{Proceedings of the 2024 on Innovation and Technology in Computer Science Education V. 1}} (Milan, Italy) \emph{(\bibinfo{series}{ITiCSE 2024})}. \bibinfo{publisher}{Association for Computing Machinery}, \bibinfo{address}{New York, NY, USA}, \bibinfo{pages}{122–128}.
\newblock
\showISBNx{9798400706004}
\urldef\tempurl%
\url{https://doi.org/10.1145/3649217.3653533}
\showDOI{\tempurl}


\bibitem[Brown et~al\mbox{.}(2020)]%
        {brown2020language}
\bibfield{author}{\bibinfo{person}{Tom Brown}, \bibinfo{person}{Benjamin Mann}, \bibinfo{person}{Nick Ryder}, \bibinfo{person}{Melanie Subbiah}, \bibinfo{person}{Jared~D Kaplan}, \bibinfo{person}{Prafulla Dhariwal}, \bibinfo{person}{Arvind Neelakantan}, \bibinfo{person}{Pranav Shyam}, \bibinfo{person}{Girish Sastry}, \bibinfo{person}{Amanda Askell}, {et~al\mbox{.}}} \bibinfo{year}{2020}\natexlab{}.
\newblock \showarticletitle{Language models are few-shot learners}.
\newblock \bibinfo{journal}{\emph{Advances in neural information processing systems}}  \bibinfo{volume}{33} (\bibinfo{year}{2020}), \bibinfo{pages}{1877--1901}.
\newblock


\bibitem[Denny et~al\mbox{.}(2023)]%
        {denny2023conversing}
\bibfield{author}{\bibinfo{person}{Paul Denny}, \bibinfo{person}{Viraj Kumar}, {and} \bibinfo{person}{Nasser Giacaman}.} \bibinfo{year}{2023}\natexlab{}.
\newblock \showarticletitle{Conversing with copilot: Exploring prompt engineering for solving cs1 problems using natural language}. In \bibinfo{booktitle}{\emph{Proc. of the 54th ACM Tech. Symp. on Computer Science Education V. 1}}. \bibinfo{pages}{1136--1142}.
\newblock


\bibitem[Denny et~al\mbox{.}(2024a)]%
        {denny2024prompt}
\bibfield{author}{\bibinfo{person}{Paul Denny}, \bibinfo{person}{Juho Leinonen}, \bibinfo{person}{James Prather}, \bibinfo{person}{Andrew Luxton-Reilly}, \bibinfo{person}{Thezyrie Amarouche}, \bibinfo{person}{Brett~A. Becker}, {and} \bibinfo{person}{Brent~N. Reeves}.} \bibinfo{year}{2024}\natexlab{a}.
\newblock \showarticletitle{Prompt Problems: A New Programming Exercise for the Generative AI Era}. In \bibinfo{booktitle}{\emph{Proceedings of the 55th ACM Technical Symposium on Computer Science Education V. 1}} (Portland, OR, USA) \emph{(\bibinfo{series}{SIGCSE 2024})}. \bibinfo{publisher}{Association for Computing Machinery}, \bibinfo{address}{New York, NY, USA}, \bibinfo{pages}{296–302}.
\newblock
\showISBNx{9798400704239}
\urldef\tempurl%
\url{https://doi.org/10.1145/3626252.3630909}
\showDOI{\tempurl}


\bibitem[Denny et~al\mbox{.}(2024b)]%
        {denny2024computing}
\bibfield{author}{\bibinfo{person}{Paul Denny}, \bibinfo{person}{James Prather}, \bibinfo{person}{Brett~A Becker}, \bibinfo{person}{James Finnie-Ansley}, \bibinfo{person}{Arto Hellas}, \bibinfo{person}{Juho Leinonen}, \bibinfo{person}{Andrew Luxton-Reilly}, \bibinfo{person}{Brent~N Reeves}, \bibinfo{person}{Eddie~Antonio Santos}, {and} \bibinfo{person}{Sami Sarsa}.} \bibinfo{year}{2024}\natexlab{b}.
\newblock \showarticletitle{Computing education in the era of generative AI}.
\newblock \bibinfo{journal}{\emph{Commun. ACM}} \bibinfo{volume}{67}, \bibinfo{number}{2} (\bibinfo{year}{2024}), \bibinfo{pages}{56--67}.
\newblock


\bibitem[Driessen et~al\mbox{.}(2024)]%
        {driessen2024putting}
\bibfield{author}{\bibinfo{person}{Tom Driessen}, \bibinfo{person}{Dimitra Dodou}, \bibinfo{person}{Pavlo Bazilinskyy}, {and} \bibinfo{person}{Joost De~Winter}.} \bibinfo{year}{2024}\natexlab{}.
\newblock \showarticletitle{Putting ChatGPT vision (GPT-4V) to the test: risk perception in traffic images}.
\newblock \bibinfo{journal}{\emph{Royal Society Open Science}} \bibinfo{volume}{11}, \bibinfo{number}{5} (\bibinfo{year}{2024}), \bibinfo{pages}{231676}.
\newblock


\bibitem[Feng et~al\mbox{.}(2024a)]%
        {feng2024more}
\bibfield{author}{\bibinfo{person}{Tony~Haoran Feng}, \bibinfo{person}{Paul Denny}, \bibinfo{person}{Burkhard~C. W\"{u}nsche}, \bibinfo{person}{Andrew Luxton-Reilly}, {and} \bibinfo{person}{Steffan Hooper}.} \bibinfo{year}{2024}\natexlab{a}.
\newblock \showarticletitle{More Than Meets the AI: Evaluating the performance of GPT-4 on Computer Graphics assessment questions}. In \bibinfo{booktitle}{\emph{Proceedings of the 26th Australasian Computing Education Conference}}. \bibinfo{pages}{182--191}.
\newblock


\bibitem[Feng et~al\mbox{.}(2024b)]%
        {feng2024can}
\bibfield{author}{\bibinfo{person}{Tony~Haoran Feng}, \bibinfo{person}{Burkhard~C. Wünsche}, \bibinfo{person}{Paul Denny}, \bibinfo{person}{Andrew Luxton-Reilly}, {and} \bibinfo{person}{Steffan Hooper}.} \bibinfo{year}{2024}\natexlab{b}.
\newblock \showarticletitle{{Can GPT-4 Trace Rays}}. In \bibinfo{booktitle}{\emph{Eurographics 2024 - Education Papers}}, \bibfield{editor}{\bibinfo{person}{Beatriz Sousa~Santos} {and} \bibinfo{person}{Eike Anderson}} (Eds.). \bibinfo{publisher}{The Eurographics Association}.
\newblock
\showISBNx{978-3-03868-238-7}
\showISSN{1017-4656}
\urldef\tempurl%
\url{https://doi.org/10.2312/eged.20241003}
\showDOI{\tempurl}


\bibitem[Finnie-Ansley et~al\mbox{.}(2022)]%
        {finnie2022robots}
\bibfield{author}{\bibinfo{person}{James Finnie-Ansley}, \bibinfo{person}{Paul Denny}, \bibinfo{person}{Brett~A Becker}, \bibinfo{person}{Andrew Luxton-Reilly}, {and} \bibinfo{person}{James Prather}.} \bibinfo{year}{2022}\natexlab{}.
\newblock \showarticletitle{The robots are coming: Exploring the implications of openai codex on introductory programming}. In \bibinfo{booktitle}{\emph{Proceedings of the 24th Australasian Computing Education Conference}}. \bibinfo{publisher}{Association for Computing Machinery}, \bibinfo{pages}{10--19}.
\newblock


\bibitem[Finnie-Ansley et~al\mbox{.}(2023)]%
        {finnie2023my}
\bibfield{author}{\bibinfo{person}{James Finnie-Ansley}, \bibinfo{person}{Paul Denny}, \bibinfo{person}{Andrew Luxton-Reilly}, \bibinfo{person}{Eddie~Antonio Santos}, \bibinfo{person}{James Prather}, {and} \bibinfo{person}{Brett~A Becker}.} \bibinfo{year}{2023}\natexlab{}.
\newblock \showarticletitle{My ai wants to know if this will be on the exam: Testing openai’s codex on cs2 programming exercises}. In \bibinfo{booktitle}{\emph{Proceedings of the 25th Australasian Computing Education Conference}}. \bibinfo{pages}{97--104}.
\newblock


\bibitem[Hirano et~al\mbox{.}(2024)]%
        {hirano2024gpt}
\bibfield{author}{\bibinfo{person}{Yuichiro Hirano}, \bibinfo{person}{Shouhei Hanaoka}, \bibinfo{person}{Takahiro Nakao}, \bibinfo{person}{Soichiro Miki}, \bibinfo{person}{Tomohiro Kikuchi}, \bibinfo{person}{Yuta Nakamura}, \bibinfo{person}{Yukihiro Nomura}, \bibinfo{person}{Takeharu Yoshikawa}, {and} \bibinfo{person}{Osamu Abe}.} \bibinfo{year}{2024}\natexlab{}.
\newblock \showarticletitle{GPT-4 Turbo with Vision fails to outperform text-only GPT-4 Turbo in the Japan Diagnostic Radiology Board Examination}.
\newblock \bibinfo{journal}{\emph{Japanese J. of Radiology}} (\bibinfo{year}{2024}), \bibinfo{pages}{1--9}.
\newblock


\bibitem[Katz et~al\mbox{.}(2024)]%
        {katz2024gpt}
\bibfield{author}{\bibinfo{person}{Daniel~Martin Katz}, \bibinfo{person}{Michael~James Bommarito}, \bibinfo{person}{Shang Gao}, {and} \bibinfo{person}{Pablo Arredondo}.} \bibinfo{year}{2024}\natexlab{}.
\newblock \showarticletitle{Gpt-4 passes the bar exam}.
\newblock \bibinfo{journal}{\emph{Philosophical Transactions of the Royal Society A}} \bibinfo{volume}{382}, \bibinfo{number}{2270} (\bibinfo{year}{2024}), \bibinfo{pages}{20230254}.
\newblock


\bibitem[Kojima et~al\mbox{.}(2022)]%
        {kojima2022large}
\bibfield{author}{\bibinfo{person}{Takeshi Kojima}, \bibinfo{person}{Shixiang~Shane Gu}, \bibinfo{person}{Machel Reid}, \bibinfo{person}{Yutaka Matsuo}, {and} \bibinfo{person}{Yusuke Iwasawa}.} \bibinfo{year}{2022}\natexlab{}.
\newblock \showarticletitle{Large language models are zero-shot reasoners}.
\newblock \bibinfo{journal}{\emph{Advances in neural information processing systems}}  \bibinfo{volume}{35} (\bibinfo{year}{2022}), \bibinfo{pages}{22199--22213}.
\newblock


\bibitem[Kumar et~al\mbox{.}(2024)]%
        {kumar2024supporting}
\bibfield{author}{\bibinfo{person}{Harsh Kumar}, \bibinfo{person}{Ruiwei Xiao}, \bibinfo{person}{Benjamin Lawson}, \bibinfo{person}{Ilya Musabirov}, \bibinfo{person}{Jiakai Shi}, \bibinfo{person}{Xinyuan Wang}, \bibinfo{person}{Huayin Luo}, \bibinfo{person}{Joseph~Jay Williams}, \bibinfo{person}{Anna~N Rafferty}, \bibinfo{person}{John Stamper}, {et~al\mbox{.}}} \bibinfo{year}{2024}\natexlab{}.
\newblock \showarticletitle{Supporting Self-Reflection at Scale with Large Language Models: Insights from Randomized Field Experiments in Classrooms}. In \bibinfo{booktitle}{\emph{Proceedings of the Eleventh ACM Conference on Learning@ Scale}}. \bibinfo{pages}{86--97}.
\newblock


\bibitem[Leinonen et~al\mbox{.}(2023)]%
        {leinonen2023using}
\bibfield{author}{\bibinfo{person}{Juho Leinonen}, \bibinfo{person}{Arto Hellas}, \bibinfo{person}{Sami Sarsa}, \bibinfo{person}{Brent Reeves}, \bibinfo{person}{Paul Denny}, \bibinfo{person}{James Prather}, {and} \bibinfo{person}{Brett~A Becker}.} \bibinfo{year}{2023}\natexlab{}.
\newblock \showarticletitle{Using large language models to enhance programming error messages}. In \bibinfo{booktitle}{\emph{Proceedings of the 54th ACM Technical Symposium on Computer Science Education V. 1}}. \bibinfo{pages}{563--569}.
\newblock


\bibitem[Li{\'e}vin et~al\mbox{.}(2024)]%
        {lievin2024can}
\bibfield{author}{\bibinfo{person}{Valentin Li{\'e}vin}, \bibinfo{person}{Christoffer~Egeberg Hother}, \bibinfo{person}{Andreas~Geert Motzfeldt}, {and} \bibinfo{person}{Ole Winther}.} \bibinfo{year}{2024}\natexlab{}.
\newblock \showarticletitle{Can large language models reason about medical questions?}
\newblock \bibinfo{journal}{\emph{Patterns}} \bibinfo{volume}{5}, \bibinfo{number}{3} (\bibinfo{year}{2024}).
\newblock


\bibitem[Liffiton et~al\mbox{.}(2023)]%
        {liffiton2023codehelp}
\bibfield{author}{\bibinfo{person}{Mark Liffiton}, \bibinfo{person}{Brad~E Sheese}, \bibinfo{person}{Jaromir Savelka}, {and} \bibinfo{person}{Paul Denny}.} \bibinfo{year}{2023}\natexlab{}.
\newblock \showarticletitle{Codehelp: Using large language models with guardrails for scalable support in programming classes}. In \bibinfo{booktitle}{\emph{Proc. of the 23rd Koli Calling Int. Conf. on Computing Education Research}}. \bibinfo{pages}{1--11}.
\newblock


\bibitem[MacNeil et~al\mbox{.}(2023)]%
        {macneil2023experiences}
\bibfield{author}{\bibinfo{person}{Stephen MacNeil}, \bibinfo{person}{Andrew Tran}, \bibinfo{person}{Arto Hellas}, \bibinfo{person}{Joanne Kim}, \bibinfo{person}{Sami Sarsa}, \bibinfo{person}{Paul Denny}, \bibinfo{person}{Seth Bernstein}, {and} \bibinfo{person}{Juho Leinonen}.} \bibinfo{year}{2023}\natexlab{}.
\newblock \showarticletitle{Experiences from using code explanations generated by large language models in a web software development e-book}. In \bibinfo{booktitle}{\emph{Proceedings of the 54th ACM Technical Symposium on Computer Science Education V. 1}}. \bibinfo{pages}{931--937}.
\newblock


\bibitem[Nori et~al\mbox{.}(2023)]%
        {nori2023capabilities}
\bibfield{author}{\bibinfo{person}{Harsha Nori}, \bibinfo{person}{Nicholas King}, \bibinfo{person}{Scott~Mayer McKinney}, \bibinfo{person}{Dean Carignan}, {and} \bibinfo{person}{Eric Horvitz}.} \bibinfo{year}{2023}\natexlab{}.
\newblock \showarticletitle{Capabilities of gpt-4 on medical challenge problems}.
\newblock \bibinfo{journal}{\emph{arXiv preprint arXiv:2303.13375}} (\bibinfo{year}{2023}).
\newblock


\bibitem[{OpenAI}(2024a)]%
        {gpt4v}
\bibfield{author}{\bibinfo{person}{{OpenAI}}.} \bibinfo{year}{2024}\natexlab{a}.
\newblock \bibinfo{title}{{GPTV\_System\_Card.pdf}}.
\newblock \bibinfo{howpublished}{\url{https://cdn.openai.com/papers/GPTV_System_Card.pdf}}.
\newblock
\newblock
\shownote{[Accessed 25-04-2024]}.


\bibitem[{OpenAI}(2024b)]%
        {gpt4o}
\bibfield{author}{\bibinfo{person}{{OpenAI}}.} \bibinfo{year}{2024}\natexlab{b}.
\newblock \bibinfo{title}{{Hello GPT-4o | OpenAI}}.
\newblock \bibinfo{howpublished}{\url{https://cdn.openai.com/papers/GPTV_System_Card.pdf}}.
\newblock
\newblock
\shownote{[Accessed 31-07-2024]}.


\bibitem[Pursnani et~al\mbox{.}(2023)]%
        {pursnani2023performance}
\bibfield{author}{\bibinfo{person}{Vinay Pursnani}, \bibinfo{person}{Yusuf Sermet}, \bibinfo{person}{Musa Kurt}, {and} \bibinfo{person}{Ibrahim Demir}.} \bibinfo{year}{2023}\natexlab{}.
\newblock \showarticletitle{Performance of ChatGPT on the US fundamentals of engineering exam: Comprehensive assessment of proficiency and potential implications for professional environmental engineering practice}.
\newblock \bibinfo{journal}{\emph{Computers and Education: Artificial Intelligence}}  \bibinfo{volume}{5} (\bibinfo{year}{2023}), \bibinfo{pages}{100183}.
\newblock


\bibitem[Reeves et~al\mbox{.}(2023)]%
        {reeves2023evaluating}
\bibfield{author}{\bibinfo{person}{Brent Reeves}, \bibinfo{person}{Sami Sarsa}, \bibinfo{person}{James Prather}, \bibinfo{person}{Paul Denny}, \bibinfo{person}{Brett~A Becker}, \bibinfo{person}{Arto Hellas}, \bibinfo{person}{Bailey Kimmel}, \bibinfo{person}{Garrett Powell}, {and} \bibinfo{person}{Juho Leinonen}.} \bibinfo{year}{2023}\natexlab{}.
\newblock \showarticletitle{Evaluating the performance of code generation models for solving Parsons problems with small prompt variations}. In \bibinfo{booktitle}{\emph{Proc. of the 2023 Conf. on Innovation and Tech. in CS Education V. 1}}. \bibinfo{pages}{299--305}.
\newblock


\bibitem[Rodrigues et~al\mbox{.}(2021)]%
        {rodrigues2021computer}
\bibfield{author}{\bibinfo{person}{Rui Rodrigues}, \bibinfo{person}{Teresa Matos}, \bibinfo{person}{Alexandre~Valle de Carvalho}, \bibinfo{person}{Jorge~G Barbosa}, \bibinfo{person}{Rodrigo Assaf}, \bibinfo{person}{Rui N{\'o}brega}, \bibinfo{person}{Ant{\'o}nio Coelho}, {and} \bibinfo{person}{A~Augusto de Sousa}.} \bibinfo{year}{2021}\natexlab{}.
\newblock \showarticletitle{Computer Graphics teaching challenges: Guidelines for balancing depth, complexity and mentoring in a confinement context}.
\newblock \bibinfo{journal}{\emph{Graphics and Visual Computing}}  \bibinfo{volume}{4} (\bibinfo{year}{2021}), \bibinfo{pages}{200021}.
\newblock


\bibitem[Savelka et~al\mbox{.}(2023)]%
        {savelka2023thrilled}
\bibfield{author}{\bibinfo{person}{Jaromir Savelka}, \bibinfo{person}{Arav Agarwal}, \bibinfo{person}{Marshall An}, \bibinfo{person}{Chris Bogart}, {and} \bibinfo{person}{Majd Sakr}.} \bibinfo{year}{2023}\natexlab{}.
\newblock \showarticletitle{Thrilled by your progress! Large language models (GPT-4) no longer struggle to pass assessments in higher education programming courses}. In \bibinfo{booktitle}{\emph{Proc. of the 2023 ACM Conf. on International Computing Education Research-Volume 1}}. \bibinfo{pages}{78--92}.
\newblock


\bibitem[Singla(2023)]%
        {singla2023evaluating}
\bibfield{author}{\bibinfo{person}{Adish Singla}.} \bibinfo{year}{2023}\natexlab{}.
\newblock \showarticletitle{Evaluating ChatGPT and GPT-4 for Visual Programming}. In \bibinfo{booktitle}{\emph{Proceedings of the 2023 ACM Conference on International Computing Education Research-Volume 2}}. \bibinfo{pages}{14--15}.
\newblock


\bibitem[Suselo et~al\mbox{.}(2017)]%
        {suselo2017journey}
\bibfield{author}{\bibinfo{person}{Thomas Suselo}, \bibinfo{person}{Burkhard~C. W{\"u}nsche}, {and} \bibinfo{person}{Andrew Luxton-Reilly}.} \bibinfo{year}{2017}\natexlab{}.
\newblock \showarticletitle{The journey to improve teaching computer graphics: A systematic review}. In \bibinfo{booktitle}{\emph{Proceedings of the 25th International Conference on Computers in Education (ICCE 2017). APSCE, Christchurch, New Zealand}}. \bibinfo{pages}{361--366}.
\newblock


\bibitem[Tu et~al\mbox{.}(2023)]%
        {tu2023should}
\bibfield{author}{\bibinfo{person}{Xinming Tu}, \bibinfo{person}{James Zou}, \bibinfo{person}{Weijie~J Su}, {and} \bibinfo{person}{Linjun Zhang}.} \bibinfo{year}{2023}\natexlab{}.
\newblock \showarticletitle{What Should Data Science Education Do with Large Language Models?}
\newblock \bibinfo{journal}{\emph{arXiv preprint arXiv:2307.02792}} (\bibinfo{year}{2023}).
\newblock


\bibitem[Wen et~al\mbox{.}(2024)]%
        {wen2024road}
\bibfield{author}{\bibinfo{person}{Licheng Wen}, \bibinfo{person}{Xuemeng Yang}, \bibinfo{person}{Daocheng Fu}, \bibinfo{person}{Xiaofeng Wang}, \bibinfo{person}{Pinlong Cai}, \bibinfo{person}{Xin Li}, \bibinfo{person}{MA Tao}, \bibinfo{person}{Yingxuan Li}, \bibinfo{person}{XU Linran}, \bibinfo{person}{Dengke Shang}, {et~al\mbox{.}}} \bibinfo{year}{2024}\natexlab{}.
\newblock \showarticletitle{On the Road with GPT-4V (ision): Explorations of Utilizing Visual-Language Model as Autonomous Driving Agent}. In \bibinfo{booktitle}{\emph{ICLR 2024 Workshop on Large Language Model (LLM) Agents}}.
\newblock


\bibitem[Wu et~al\mbox{.}(2023a)]%
        {wu2023can}
\bibfield{author}{\bibinfo{person}{Chaoyi Wu}, \bibinfo{person}{Jiayu Lei}, \bibinfo{person}{Qiaoyu Zheng}, \bibinfo{person}{Weike Zhao}, \bibinfo{person}{Weixiong Lin}, \bibinfo{person}{Xiaoman Zhang}, \bibinfo{person}{Xiao Zhou}, \bibinfo{person}{Ziheng Zhao}, \bibinfo{person}{Ya Zhang}, \bibinfo{person}{Yanfeng Wang}, {et~al\mbox{.}}} \bibinfo{year}{2023}\natexlab{a}.
\newblock \showarticletitle{Can gpt-4v (ision) serve medical applications? case studies on gpt-4v for multimodal medical diagnosis}.
\newblock \bibinfo{journal}{\emph{arXiv preprint arXiv:2310.09909}} (\bibinfo{year}{2023}).
\newblock


\bibitem[Wu et~al\mbox{.}(2023b)]%
        {wu2023early}
\bibfield{author}{\bibinfo{person}{Yang Wu}, \bibinfo{person}{Shilong Wang}, \bibinfo{person}{Hao Yang}, \bibinfo{person}{Tian Zheng}, \bibinfo{person}{Hongbo Zhang}, \bibinfo{person}{Yanyan Zhao}, {and} \bibinfo{person}{Bing Qin}.} \bibinfo{year}{2023}\natexlab{b}.
\newblock \showarticletitle{An early evaluation of gpt-4v (ision)}.
\newblock \bibinfo{journal}{\emph{arXiv preprint arXiv:2310.16534}} (\bibinfo{year}{2023}).
\newblock


\bibitem[W{\"u}nsche et~al\mbox{.}(2018)]%
        {wunsche2018automatic}
\bibfield{author}{\bibinfo{person}{Burkhard~C. W{\"u}nsche}, \bibinfo{person}{Zhen Chen}, \bibinfo{person}{Lindsay Shaw}, \bibinfo{person}{Thomas Suselo}, \bibinfo{person}{Kai-Cheung Leung}, \bibinfo{person}{Davis Dimalen}, \bibinfo{person}{Wannes van~der Mark}, \bibinfo{person}{Andrew Luxton-Reilly}, {and} \bibinfo{person}{Richard Lobb}.} \bibinfo{year}{2018}\natexlab{}.
\newblock \showarticletitle{Automatic assessment of OpenGL computer graphics assignments}. In \bibinfo{booktitle}{\emph{Proceedings of the 23rd annual ACM conference on innovation and technology in computer science education}}. \bibinfo{pages}{81--86}.
\newblock


\bibitem[W\"{u}nsche et~al\mbox{.}(2019)]%
        {wunsche2019automatic}
\bibfield{author}{\bibinfo{person}{Burkhard~C. W\"{u}nsche}, \bibinfo{person}{Edward Huang}, \bibinfo{person}{Lindsay Shaw}, \bibinfo{person}{Thomas Suselo}, \bibinfo{person}{Kai-Cheung Leung}, \bibinfo{person}{Davis Dimalen}, \bibinfo{person}{Wannes van~der Mark}, \bibinfo{person}{Andrew Luxton-Reilly}, {and} \bibinfo{person}{Richard Lobb}.} \bibinfo{year}{2019}\natexlab{}.
\newblock \showarticletitle{CodeRunnerGL - An Interactive Web-Based Tool for Computer Graphics Teaching and Assessment}. In \bibinfo{booktitle}{\emph{Proceedings of the International Conference on Electronics, Information, and Communication {(ICEIC 2019)}}}. \bibinfo{publisher}{IEEE}, \bibinfo{address}{New York, NY, USA}, \bibinfo{pages}{1--7}.
\newblock
\urldef\tempurl%
\url{https://doi.org/10.23919/ELINFOCOM.2019.8706402}
\showDOI{\tempurl}


\bibitem[Xu and Tao(2024)]%
        {xu2024map}
\bibfield{author}{\bibinfo{person}{Jinwen Xu} {and} \bibinfo{person}{Ran Tao}.} \bibinfo{year}{2024}\natexlab{}.
\newblock \showarticletitle{Map Reading and Analysis with GPT-4V (ision)}.
\newblock \bibinfo{journal}{\emph{ISPRS International Journal of Geo-Information}} \bibinfo{volume}{13}, \bibinfo{number}{4} (\bibinfo{year}{2024}), \bibinfo{pages}{127}.
\newblock


\bibitem[Yang et~al\mbox{.}(2022)]%
        {yang2022empirical}
\bibfield{author}{\bibinfo{person}{Zhengyuan Yang}, \bibinfo{person}{Zhe Gan}, \bibinfo{person}{Jianfeng Wang}, \bibinfo{person}{Xiaowei Hu}, \bibinfo{person}{Yumao Lu}, \bibinfo{person}{Zicheng Liu}, {and} \bibinfo{person}{Lijuan Wang}.} \bibinfo{year}{2022}\natexlab{}.
\newblock \showarticletitle{An empirical study of gpt-3 for few-shot knowledge-based vqa}. In \bibinfo{booktitle}{\emph{Proceedings of the AAAI conference on artificial intelligence}}, Vol.~\bibinfo{volume}{36}. \bibinfo{pages}{3081--3089}.
\newblock


\bibitem[Yang et~al\mbox{.}(2023)]%
        {yang2023dawn}
\bibfield{author}{\bibinfo{person}{Zhengyuan Yang}, \bibinfo{person}{Linjie Li}, \bibinfo{person}{Kevin Lin}, \bibinfo{person}{Jianfeng Wang}, \bibinfo{person}{Chung-Ching Lin}, \bibinfo{person}{Zicheng Liu}, {and} \bibinfo{person}{Lijuan Wang}.} \bibinfo{year}{2023}\natexlab{}.
\newblock \showarticletitle{The dawn of lmms: Preliminary explorations with gpt-4v (ision)}.
\newblock \bibinfo{journal}{\emph{arXiv preprint arXiv:2309.17421}} \bibinfo{volume}{9}, \bibinfo{number}{1} (\bibinfo{year}{2023}), \bibinfo{pages}{1}.
\newblock


\bibitem[Yeadon and Hardy(2023)]%
        {yeadon2023impact}
\bibfield{author}{\bibinfo{person}{Will Yeadon} {and} \bibinfo{person}{Tom Hardy}.} \bibinfo{year}{2023}\natexlab{}.
\newblock \showarticletitle{The Impact of AI in Physics Education: A Comprehensive Review from GCSE to University Levels}.
\newblock \bibinfo{journal}{\emph{arXiv preprint arXiv:2309.05163}} (\bibinfo{year}{2023}).
\newblock


\bibitem[Zain et~al\mbox{.}(2023)]%
        {zain2023use}
\bibfield{author}{\bibinfo{person}{Iffah~NM Zain}, \bibinfo{person}{Mohd~AB Setambah}, \bibinfo{person}{Mohd~S Othman}, {and} \bibinfo{person}{Mazarul~HM Hanapi}.} \bibinfo{year}{2023}\natexlab{}.
\newblock \showarticletitle{Use of Photomath Applications in Helping Improving Students’ Mathematical (Algebra) Achievement}.
\newblock \bibinfo{journal}{\emph{European Journal of Education and Pedagogy}} \bibinfo{volume}{4}, \bibinfo{number}{2} (\bibinfo{year}{2023}), \bibinfo{pages}{85--87}.
\newblock


\end{thebibliography}
\end{document}